# Graph Neural AI with Temporal Dynamics for Comprehensive Anomaly Detection in Microservices

Qingyuan Zhang
Boston University
Boston, USA

Ning Lyu
Carnegie Mellon University
Pittsburgh, USA

Le Liu
University of California, San Diego
San Diego, USA

Yuxi Wang
Hofstra University
Hempstead, USA

Ziyu Cheng
University of Southern California
Los Angeles, USA

Cancan Hua*
University of Southern California
Los Angeles, USA

*Abstract-This study addresses the problem of anomaly detection and root cause tracing in microservice architectures and proposes a unified framework that combines graph neural networks with temporal modeling. The microservice call chain is abstracted as a directed graph, where multidimensional features of nodes and edges are used to construct a service topology representation, and graph convolution is applied to aggregate features across nodes and model dependencies, capturing complex structural relationships among services. On this basis, gated recurrent units are introduced to model the temporal evolution of call chains, and multi-layer stacking and concatenation operations are used to jointly obtain structural and temporal representations, improving the ability to identify anomaly patterns. Furthermore, anomaly scoring functions at both the node and path levels are defined to achieve unified modeling from local anomaly detection to global call chain tracing, which enables the identification of abnormal service nodes and the reconstruction of potential anomaly propagation paths. Sensitivity experiments are then designed from multiple dimensions, including hyperparameters, environmental disturbances, and data distribution, to evaluate the framework, and results show that it outperforms baseline methods in key metrics such as AUC, ACC, Recall, and F1-Score, maintaining high accuracy and stability under dynamic topologies and complex environments. This research not only provides a new technical path for anomaly detection in microservices but also lays a methodological foundation for intelligent operations in distributed systems.*

## I. INTRODUCTION

In modern internet systems, microservice architecture has gradually become mainstream. By decomposing large and complex monolithic applications into a set of loosely coupled independent services, it enables better scalability and flexibility. However, as the number of services and the complexity of interactions increase, abnormal phenomena during system operation occur more frequently[1]. The call chain, as a key reflection of microservice runtime states, records the full request path and service dependencies. Once an anomaly occurs, it may not only reduce the performance of a single service but also trigger cascading effects in cross-service dependencies, ultimately threatening system stability and availability. Therefore, achieving efficient anomaly detection and root cause tracing at the call chain level has become a critical issue in backend architecture governance[2].

Against this background, anomaly detection has become increasingly important. Traditional single-point monitoring or rule-based methods often fail to address the high-dimensional and multimodal data characteristics of complex distributed systems. On one hand, the number of request paths in a microservice call chain is enormous, and the topology changes dynamically. Anomalies may no longer be limited to performance indicators of a single node but may appear as structural patterns or temporal dependencies. On the other hand, the uncertainty and heterogeneity of the system environment further increase modeling difficulty. If backend anomalies cannot be detected and localized in time, they directly affect user experience and may cause severe economic losses and trust crises. These practical demands drive research into global anomaly detection using intelligent algorithms[3].

Compared with traditional approaches, graph structures provide a natural representation of microservice call chains. Essentially, a call chain is a directed graph composed of service nodes and invocation edges, with topology and edge weights carrying rich interaction information. Graph Neural Networks (GNNs) have been extensively applied across a wide range of domains, including medical diagnosis [4-6], financial risk assessment [7] and fraud detection [8], as well as enhancing reasoning and knowledge representation in large language models [9-10]. In this context, graph neural networks offer a more suitable modeling tool. By embedding and propagating node and edge features, they capture inter-service correlations and contextual dependencies, revealing hidden abnormal patterns in complex topologies. Graph-based modeling not only overcomes the limitations of traditional statistical methods but also preserves structural information under dynamic invocation relationships, offering new directions for anomaly detection and root cause tracing[11].

The study of anomaly detection in microservice call chains is significant not only for improving system stability but also for promoting intelligent transformation in backend service governance. With advanced models such as graph neural networks, anomalies can be automatically identified, accurately localized, and explained[12]. This substantially reduces the cost and time of manual troubleshooting. For large internet enterprises, financial trading platforms, and critical infrastructure systems, such methods hold high value. They help ensure stable operation under high concurrency and uncertain environments. They also provide a solid foundation for building adaptive service scheduling and fault-tolerant mechanisms. As system complexity continues to grow, anomaly detection will no longer be a single function but an indispensable core capability in backend architectures[13].

Furthermore, from the perspective of integrating academic research with industrial practice, this direction also supports the development of intelligent operations and maintenance. Backend systems are moving from passive monitoring to active diagnosis, and further toward prediction and prevention. Research on anomaly detection and root cause tracing in microservice call chains not only safeguards existing system stability but also serves as a prerequisite for intelligent backend governance [14]. It contributes to refining theoretical systems of anomaly detection and helps enterprises build reliable, interpretable, and sustainable service ecosystems. With the ongoing convergence of artificial intelligence and distributed systems, this research area will unleash broader application potential and drive backend architectures toward greater intelligence and autonomy.

## II. PROPOSED APPROACH

In designing the method, we first need to formally represent the microservice call chain. The system can be abstracted as a directed graph $G = (V, E)$, where V represents the set of service nodes and E represents the set of call edges. Each edge is accompanied by weight information to characterize call frequency or latency. In this graph structure, the feature vector of each service node is denoted as $x_i \in R^d$, and the overall node feature matrix is $X \in R^{|V| \times d}$. The topology of the call chain is described by the adjacency matrix $A \in R^{|V| \times |V|}$, where $A_{ij}$ represents the call relationship between node i and node j. Based on this modeling approach, we employ graph convolution to aggregate and propagate cross-service features, enabling the model to effectively capture global dependencies in the microservice call chain. Drawing on the self-supervised transfer learning strategies introduced by Zhou, our framework applies shared encoder designs to enhance knowledge transfer and improve the model's adaptability across diverse cloud service domains [15]. We also integrate contrastive learning-based dependency modeling as proposed by Xing et al., strengthening the representation of inter-service relationships and enabling more accurate identification of anomalous dependency patterns [16]. To ensure both efficiency and data privacy in large-scale distributed environments, the architecture further leverages federated learning techniques described by Liu et al., supporting privacy-preserving feature aggregation and minimizing communication overhead during model training [17]. The overall model structure, incorporating these methodological advances, is illustrated in Figure 1.

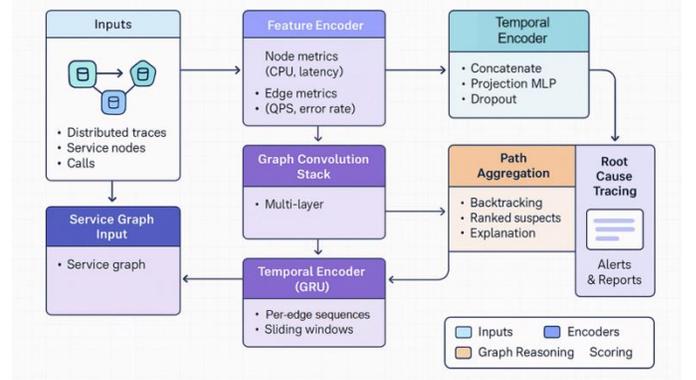

Figure 1. Overall model architecture

In the propagation mechanism of graph neural networks, the update of node representation can be written as follows:

$$H^{(l+1)} = \sigma(\widetilde{D}^{-\frac{1}{2}} \widetilde{A} \widetilde{D}^{-\frac{1}{2}} H^{(l)} W^{(l)}) \qquad (1)$$

Here, $\widetilde{A} = A + I$ represents the adjacency matrix after adding a self-connection, $\widetilde{D}$ is the degree matrix of $\widetilde{A}$, $H^{(l)}$ represents the node embeddings at layer 1, $W^{(l)}$ is the learnable parameter matrix, and $\sigma(\cdot)$ represents the nonlinear activation function. By stacking multiple layers, the receptive field of node representations can be gradually expanded, capturing longer-range call dependencies. Furthermore, to avoid oversmoothing, residual connections and layer normalization can be introduced to maintain the stability of deep networks.

To enhance the characterization of dynamic changes in call chains, a temporal modeling mechanism is introduced. Let the temporal feature of each call edge be $z_{ij}(t)$, which can be modeled using a gated recurrent unit:

$$h_t = GRU(z_{ij}(t), h_{t-1}) \qquad (2)$$

Where $h_t$ represents the hidden state at time t. This structure can effectively capture the dependency patterns of the call chain that evolve and combine them with graph structure information for multimodal feature fusion. Ultimately, the node representation is composed of the structural features and temporal features, denoted as:

$$u_i = [h_i^{struct} \| h_i^{temp}] \qquad (3)$$

Among them, $h_i^{struct}$ comes from the output of graph convolution, $h_i^{temp}$ comes from the output of temporal modeling, and the symbol $\|$ represents the vector concatenation operation.

In anomaly detection and root cause tracing, it is necessary to construct a discriminant function to measure the degree of abnormality of a node or subgraph. By defining the score function:

$$s_i = \| u_i - \mu \|_2^2 \quad (4)$$

Where $\mu$ represents the central embedding vector of all nodes, and $s_i$ reflects the degree to which node i deviates from the normal pattern. Furthermore, aggregation operations can be introduced at the subgraph level to perform an overall score on the call path:

$$S(P) = \frac{1}{|P|} \sum_{i \in P} s_i \quad (5)$$

$P$ represents a path in the call chain, and $S(P)$ reflects the overall abnormality level of that path. This approach enables joint node-level and path-level analysis, enabling both the identification of abnormal service nodes and the tracing of potential root cause call chains, thus supporting the stability of backend systems.

III. PERFORMANCE EVALUATION

*A. Dataset*

This study adopts the DeathStarBench (Social Network) distributed tracing dataset as a representative data source for microservice call chain scenarios. The dataset is derived from a multilingual microservice application designed for social network services. It includes dozens of independent services and complex invocation dependencies, covering typical functions such as user homepage, content publishing, relationship queries, media processing, and URL shortening. The system runs under realistic service decomposition and a no-shared-database mode. It generates logged cross-service request traces and performance variations under high concurrency, which closely reproduce the structural and operational characteristics of modern internet backends.

The data are organized in an OpenTracing/Jaeger-compatible span record format. Each record contains key fields such as trace_id, span_id, parent_span_id, service_name, operation_name, start_ts, duration, tags, and logs. These fields capture the end-to-end path of a request through the microservice network. Based on parent-child relationships, the call tree can be reconstructed and aggregated within a time window to obtain the service graph. Node features can be derived from statistics such as latency, error rate, throughput, and resource usage. Edge features can be derived from metrics such as call frequency, retry counts, and timeout counts. This information reflects both the topological structure and the temporal context, enabling joint modeling of structure and time.

The dataset is chosen for its high realism and high observability. On one hand, the large number of services, deep call chains, and wide fan-out produce a dynamic topology that changes with workload and function composition. This makes it naturally suitable for verifying anomaly patterns and propagation effects at the call chain level. On the other hand, the tracing records contain complete fields, which allow direct construction of service graphs and call path sets from raw spans. These can be aligned with time series metrics to support multi-granularity analysis in anomaly detection and root cause tracing. In addition, the dataset does not involve personal sensitive information, making it easy to reproduce and share. It provides a stable, public, and engineering-oriented benchmark for evaluating the "graph plus time series" paradigm.

*B. Experimental Results*

This paper first conducts a comparative experiment, and the experimental results are shown in Table 1.

Table 1. Comparative experimental results

| Method | AUC | ACC | Recall | F1-Score |
|---|---|---|---|---|
| 1DCNN[18] | 0.873 | 0.846 | 0.821 | 0.832 |
| LSTM[19] | 0.889 | 0.854 | 0.837 | 0.845 |
| Transformer[20] | 0.913 | 0.872 | 0.856 | 0.864 |
| GAT[21] | 0.928 | 0.881 | 0.867 | 0.873 |
| Ours | 0.951 | 0.904 | 0.889 | 0.896 |

From the overall performance, the results in Table 1 show that deep modeling methods have clear advantages over traditional one-dimensional convolution and recurrent networks. Although 1DCNN and LSTM have some ability to capture temporal patterns, they cannot fully model the complex cross-node dependencies in microservice call chains. Their performance on most metrics remains relatively low. In particular, for AUC and F1-Score, these methods rely more on local features, which limits their capability to identify anomalies in complex scenarios.

Further observations reveal that the Transformer outperforms LSTM and 1DCNN across all four core metrics. This indicates that its self-attention mechanism can better capture global dependencies and model associations in long sequences. Such capability is especially important for multi-hop calls and dynamic topologies in microservice call chains. As a result, both ACC and Recall are improved. However, relying only on temporal attention is not sufficient to fully exploit the graph structural characteristics of call chains, and limitations remain in root cause tracing.

In comparison, the GAT achieves further performance improvements. This suggests that treating the call chain as a graph and incorporating graph attention mechanisms can effectively enhance the accuracy of anomaly detection and localization. By propagating information between nodes and edges, the model can better capture both local and global structural patterns, which leads to advantages in Recall and F1-Score. These results highlight the value of call chain topology in anomaly detection and demonstrate the importance of structural modeling for complex distributed systems.

Finally, the proposed method achieves the best results across all metrics, with significant gains in both AUC and ACC. This shows that the graph neural network and temporal joint modeling framework can capture structural relationships among services while integrating dynamic temporal evolution features. It provides higher accuracy and stability in both anomaly detection and root cause localization. These advantages confirm the effectiveness of the method in microservice call chain scenarios and offer strong support for intelligent operations and maintenance in backend systems.

This paper also presents a single-factor sensitivity experiment on the weight decay coefficient to F1-Score, and the experimental results are shown in Figure 2.

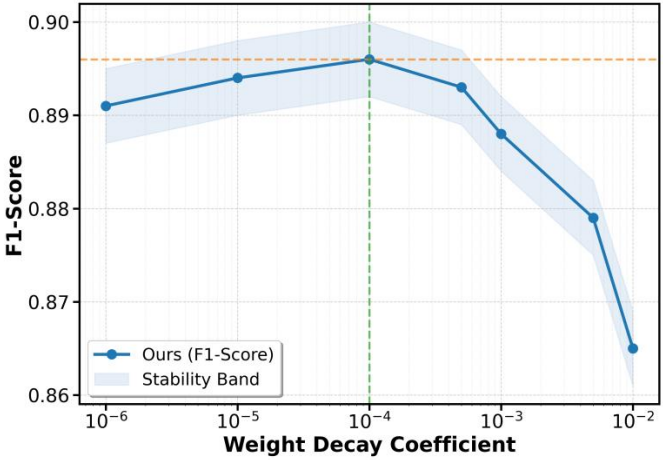

Figure 2. Single-factor sensitivity experiment of weight decay coefficient to F1-Score

From the figure, it can be observed that as the weight decay coefficient increases, the F1-Score first shows an upward trend and reaches its peak around $10^{-4}$. This indicates that with moderate regularization, the model can effectively suppress overfitting. As a result, anomaly detection becomes more robust and achieves higher accuracy and recall in complex call chain scenarios.

When the weight decay coefficient exceeds the optimal point, the F1-Score drops significantly. This result shows that excessive regularization weakens the model's ability to represent call chain structures and temporal features. Valuable patterns are over-smoothed, reducing the model's ability to distinguish anomalies. This effect is particularly evident in distributed systems, where dependencies between services contain high uncertainty and fluctuations. Over-constraining reduces the model's sensitivity.

The stable region shows that the model exhibits small performance fluctuations within the range of $10^{-5}$ to $10^{-4}$. It demonstrates strong robustness. This means that within this parameter range, the method balances generalization and fitting, ensuring stable anomaly detection and root cause tracing under different workload conditions. For real systems, such stability is crucial, as call patterns and resource pressures in backends often change dynamically.

Overall, this sensitivity experiment confirms the importance of choosing an appropriate weight decay coefficient to guarantee model performance. Moderate regularization improves generalization and strengthens the practicality and reliability of anomaly detection frameworks in complex microservice call chains. This finding highlights the key role of parameter tuning in intelligent operations and maintenance. It also provides reference values for automated tuning and adaptive optimization in dynamic environments.

This article further presents a sensitivity experiment on the elastic scaling frequency of service instances, the results of which are shown in Figure 3.

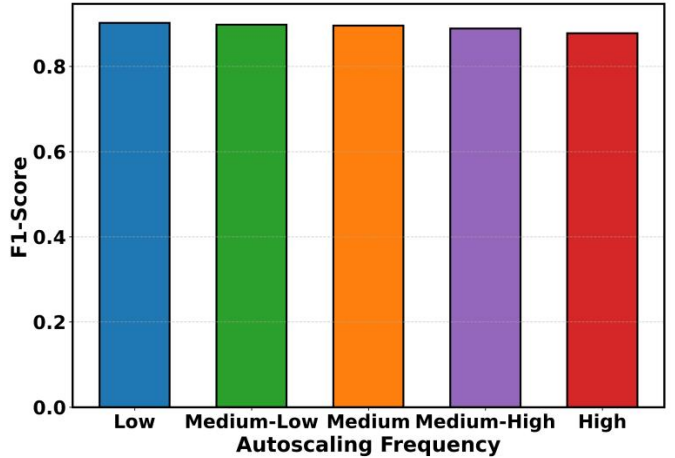

Figure 3. Sensitivity experiment of service instance elastic scaling frequency

The results show that when the frequency of service instance scaling is low or at a moderate level, the model maintains a high F1-Score. This indicates that in this range, the system topology changes are relatively stable. The model can effectively capture both structural features of the call chain and temporal dependencies, which sustains strong anomaly detection and root cause tracing capabilities. Moderate scaling provides dynamism without causing excessive disturbances, so the detection performance remains stable.

As the frequency of scaling increases, the F1-Score shows a downward trend. This suggests that frequent instance changes introduce significant disturbances to the service topology and call chain relationships, which increases the difficulty of anomaly detection. High-frequency scaling may cause continuous changes in call chain contexts, making it difficult for the model to learn and adapt to new dependency patterns in a short time. This weakens the stability of anomaly recognition. The phenomenon highlights the need for stronger robustness and adaptive capability in highly dynamic environments, so that anomaly detection frameworks can maintain reliability in real backend systems.

IV. CONCLUSION

This paper studies the problem of anomaly detection and root cause tracing in microservice call chains and proposes an optimized framework that combines graph neural networks with temporal modeling. The study begins with the structural characteristics of call chains and cross-node dependencies. A graph representation method is constructed to capture multi-hop interactions and dynamic evolution. By integrating temporal encoding mechanisms, the framework improves the accuracy and stability of anomaly detection. In multiple comparative and sensitivity experiments, the method demonstrates adaptability and robustness in complex distributed system environments, providing a new approach for backend service governance.

In terms of overall contribution, this work expands the research paradigm of microservice anomaly detection and verifies the feasibility and value of graph-based learning in service governance. By considering both structural and temporal features, the method overcomes the limitations of traditional single-metric or rule-based approaches. It maintains a strong detection capability when facing large-scale and dynamic call chains. These findings enrich the research path of intelligent operations in distributed systems and provide a modeling framework that can be referenced in related fields.

From an application perspective, the study has significant value for ensuring the stable operation of critical business systems. In high-concurrency and high-complexity scenarios such as internet services, financial transactions, smart healthcare, and logistics, system stability and rapid anomaly response are directly linked to user experience and economic benefits. The proposed method enables automated and intelligent anomaly detection and root cause analysis, which shortens troubleshooting time, reduces manual costs, and supports system resilience design and resource scheduling. This capability helps advance industry applications toward higher reliability and availability.

## V. Future Work

Looking ahead, as distributed architectures continue to evolve, the scale and complexity of microservices will keep increasing. Efficient anomaly detection and tracing in larger and more dynamic environments will remain an important challenge. Future research can incorporate adaptive regularization, multimodal feature fusion, and federated modeling to enhance generalization across domains and platforms. Integrating the framework with intelligent scheduling and automated operations platforms also holds promise for building more autonomous and intelligent backend governance systems, providing a solid guarantee for the security and stability of large-scale systems.